\documentclass{article}

\usepackage{arxiv}

\usepackage[utf8]{inputenc} 
\usepackage[T1]{fontenc}    
\usepackage[hidelinks]{hyperref}     
\usepackage{url}            
\usepackage{booktabs}       
\usepackage{amsfonts}       
\usepackage{nicefrac}       
\usepackage{microtype}      
\usepackage{xcolor}         
\usepackage{graphicx}

\title{Video Diffusion Transformers are In-Context Learners}

\author{%
Zhengcong Fei, Di Qiu, Changqian Yu\thanks{Corresponding author}, Debang Li
\\
\textbf{Mingyuan Fan, Xiang Wen} \\
Kunlun Inc.\\
Beijing, China\\
{\tt\small \{feizhengcong\}@gmail.com}
}

\begin{document}

\maketitle

\begin{abstract}
  
This paper investigates a solution for enabling in-context capabilities ofmain video diffusion transformers, with minimal tuning required for activation. Specifically, we propose a simple pipeline to leverage in-context generation: ($\textbf{i}$) concatenate videos along spacial or time dimension, ($\textbf{ii}$) jointly caption multi-scene video clips from one source, and ($\textbf{iii}$) apply task-specific fine-tuning using carefully curated small datasets. Through a series of diverse controllable tasks, we demonstrate qualitatively that existing advanced text-to-video models can effectively perform in-context generation.
Notably, it allows for the creation of consistent multi-scene videos exceeding 30 seconds in duration, without additional computational overhead.
Importantly, this method requires no modifications to the original models, results in high-fidelity video outputs that better align with prompt specifications and maintain role consistency. 
Our framework presents a valuable tool for the research community and offers critical insights for advancing product-level controllable video generation systems.
The code and model weights are publicly available.

\end{abstract}

\section{Introduction}

Diffusion models have set a new benchmark in vision generation tasks, by employing an iterative denoising process applied to a Gaussian noise latent space \cite{ho2020ddpm,song2020ddim,peebles2023dit,esser2024scaling,rombach2022high,gu2022vector,saharia2022photorealistic,ramesh2022hierarchical}. Such iterative denoising process has proven to be both robust and versatile in the context of video generation. Meantime, large-scale, pre-trained video diffusion models \cite{blattmann2023stable,girdhar2023emu,polyak2024movie,yang2024cogvideox} have greatly advanced downstream applications, with many methods providing enhanced control over various attributes, thereby enabling more precise adjustments during the generation process \cite{zhang2023adding,ruiz2023dreambooth,gal2022image,fei2023gradient}. Despite these advancements, adapting text-to-video models for a wide range of generative tasks, especially those requiring the creation of coherent video sets with complex intrinsic relationships, remains an open challenge \cite{hu2023videocontrolnet,peng2024controlnext,zhang2023controlvideo}. Moreover, the generation and control of multi-scene, long-duration videos continue to pose difficulties due to the complexity of the spatio-temporal domain and the substantial GPU memory requirements.

\begin{figure*}[htbp]
  \centering
   \includegraphics[width=0.99\linewidth]{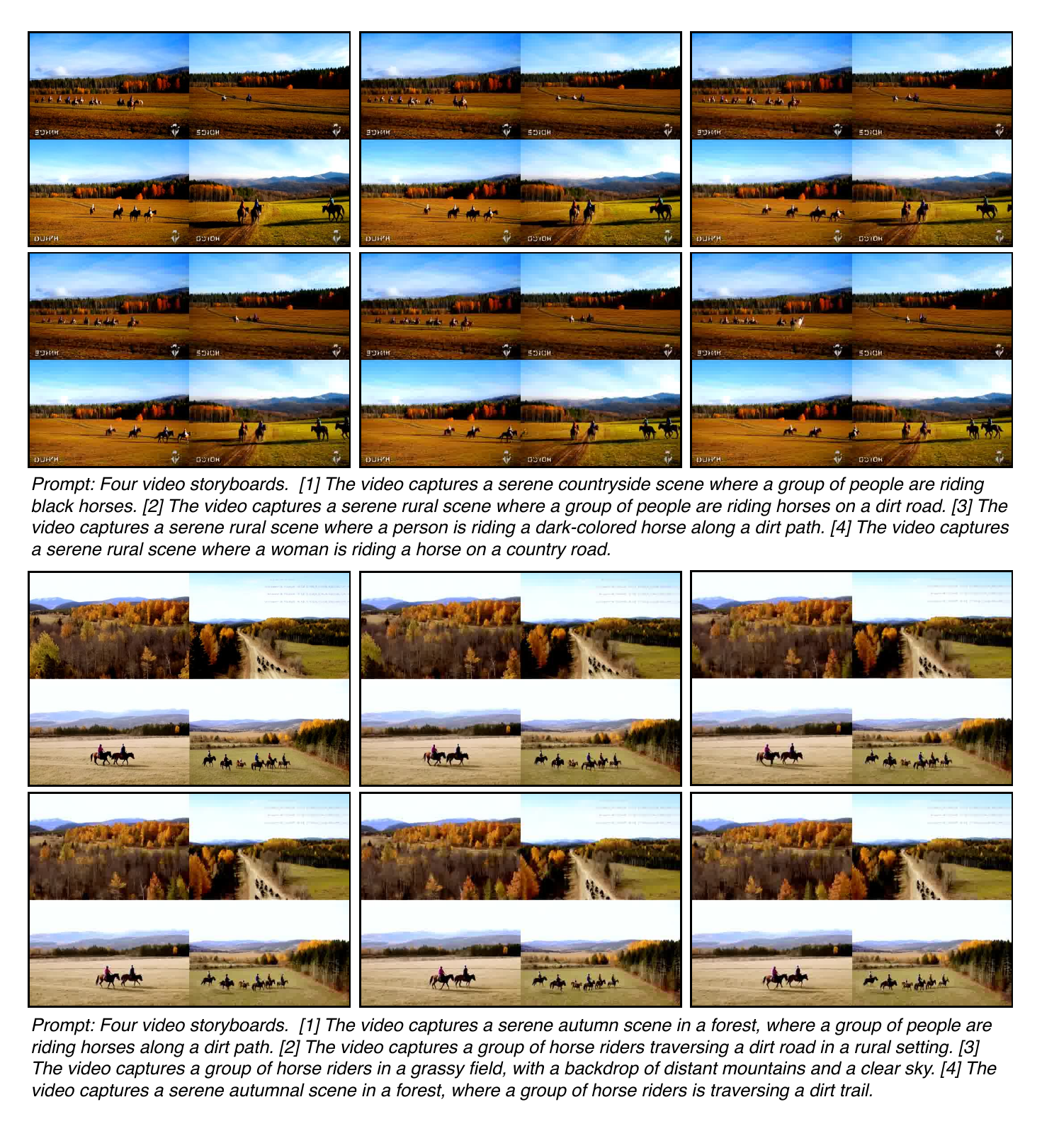}
   \caption{\textbf{Examples of in-context generalist for multi-scene video tasks.} Four sub-videos are concurrently generated within a single diffusion process that are tuned specifically. A carefully designed prompt template, incorporating distinct scenes, is employed to ensure consistent portrayal and seamless integration of scenes in the generated video sets.
   }
   \label{fig:case1} 
\end{figure*}

In-context learning (ICL), initially identified as a latent capability in large autoregressive language models \cite{radford2021learning}, has garnered significant attention in recent research \cite{dong2022survey}. This technique involves providing models with a small set of labeled data as controllable demonstrations, enhancing their ability to reason and perform specific tasks \cite{brown2020language, dong2022survey, wang2023label}. ICL has been shown to notably improve model performance across a wide range of tasks \cite{garcia2023unreasonable, moslem2023adaptive, qin2023context, li2023few}. Through ICL, models can dynamically adjust their behavior according to the contextual information provided, enabling them to execute tasks with minimal labeled data. 
The success of ICL in language models has sparked interest in extending these capabilities to vision tasks. However, implementing vision-based ICL presents greater challenges than in language models due to the inherently high-dimensional and diverse nature of visual data \cite{kirillov2019panoptic, zhang2016colorful}, particularly when dealing with video data. As the temporal dimension in videos introduces dynamic causality and long-sequence complexity.

Inspired from \cite{huang2024context}, we aim at adapting text-to-video models for conditional control tasks with in-context capacity, offering a universal approach for multi-scene long-duration video generation. 
Our approach begins with a key assumption: 
\emph{Although trained with cut video clips, text-to-video models possess in-context generation capabilities for homologous scenes in inner connection. }
By effectively triggering and enhancing this built-in functionality, we aim to leverage it for more complex conditional generative tasks. 
Specifically, we concatenate multiple video clips from one source into an extended video sequence and combine individual video prompts into one comprehensive prompt, enabling the model to process and generate multiple-scene videos simultaneously. 
Then, we fine-tune a Low-Rank Adaptation (LoRA) \cite{hu2021lora} of the model with a small set of carefully designed samples. 
It effectively reduces the computational resources required while preserving much of the original model's knowledge and in-context generation capabilities.

Despite its simplicity transferring from image generation domain, it demonstrates remarkable adaptability and high-quality generative performance across a diverse range of conditional tasks. Example outputs for multi-scene tasks that create videos exceeding 30 seconds in one shot, are illustrated in Figure \ref{fig:case1}, specific  portrait photograph and style transfer task cases are shown in Figure \ref{fig:case2}, and conditional generation results, including outpainting and inpainting, are presented in Figure \ref{fig:case3}.
Although our method necessitates small-scale, task-specific tuning data to enhance in-context capabilities for generative controllability, the core video diffusion transformer and pipeline remain task-agnostic. 
Notably, our method also can generate consistent multi-scene videos exceeding 30 seconds without additional computational overhead. 
This combination of minimal data requirements and broad applicability provides a valuable tool for the generative community. 
To support continued exploration, we have made our data, models, and training configurations publicly available on GitHub.

\section{Related Works}

\subsection{Text-to-Video Generation}

Generation of video remains a longstanding challenge in the field of computer vision \cite{ho2022imagenvideo, yu2023video}. In between, text-to-video generation has progressed significantly over time, evolving from early approaches based on GANs and VAEs to more advanced structures such as diffusion. Initial GAN-based methods \cite{vondrick2016generating, saito2017temporal, tulyakov2018mocogan, clark2019adversarial, yu2022generating} faced dilemma with temporal coherence, leading to inconsistencies between consecutive frames. To address these issues, video diffusion models based on U-Net adapted originally designed for text-to-image, thereby improving the continuity of frames. Furthermore, DiT-based architectures \cite{peebles2023dit, lu2023vdt, ma2024latte, gao2024lumina} introduced spatio-temporal attention and full 3d attention, which enhanced the ability to capture the intricate dynamics of video and ensure greater frame consistency \cite{he2022latent, blattmann2023align, chen2023seine, girdhar2023emuvideo}. Meanwhile, auto-regressive models \cite{yan2021videogpt, hong2022cogvideo, villegas2022phenaki, kondratyuk2023videopoet, xie2024show, liu2024mardini} employed token-based approaches to effectively model temporal dependencies, particularly excelling in the generation of long-form videos \cite{yin2023nuwa, wang2023genlvideo, zhao2024moviedreamer, henschel2024streamingt2v, tan2024videoinfinity, zhou2024storydiffusion} and in video-to-video translation tasks \cite{yang2023rerender, bao2023latentwarp, yatim2024smm, hu2024depthcrafter}.  Our work investigates the effectiveness of video diffusion Transformers for in-context capacity in long-duration multi-scene tasks and use a lightweight fine-tuning to activate it.

\subsection{Diffusion Transformers}

Transformer architecture \cite{vaswani2017attention} has garnered significant attention for its success in language modeling \cite{radford2018improving, radford2019language, raffel2020exploring} and substantial potential across a wide range of computer vision applications, including image classification \cite{dosovitskiy2020image, he2022masked, touvron2021training}, object detection \cite{liu2021swin, wang2021pyramid, wang2022pvt, carion2020end}, and semantic segmentation \cite{zheng2021rethinking, xie2021segformer, strudel2021segmenter}, among others \cite{liu2022video, he2022masked,fei2023masked,fei2021partially,fei2022deecap,fei2019fast,fei2022progressive}. Building on this success, diffusion Transformers \cite{Peebles_2023} and their variants \cite{bao2023all, fei2024scalable,fei2024scaling,fei2024dimba,fei2024diffusion} have replaced the traditional convolutional U-Net backbone \cite{ronneberger2015u} with Transformer-based architectures. This shift has resulted in improved scalability and simpler parameter expansion compared to U-Net-based diffusion models. Such advantages in scalability have been particularly evident in areas like video generation \cite{ma2024latte}, image generation \cite{chen2023pixart}, speech generation \cite{liu2023vit}, and video generation \cite{yang2024cogvideox,fei2025ingredients}. Furthermore, recent innovations, such as StableAudio \cite{evans2024stable} and FluxMusic \cite{fei2024flux,fei2024music,fei2023jepa,qiu2025skyreels}, have explored the application of the DiT architecture in the generation of audio and music.

\begin{figure*}[htbp]
  \centering
   \includegraphics[width=0.99\linewidth]{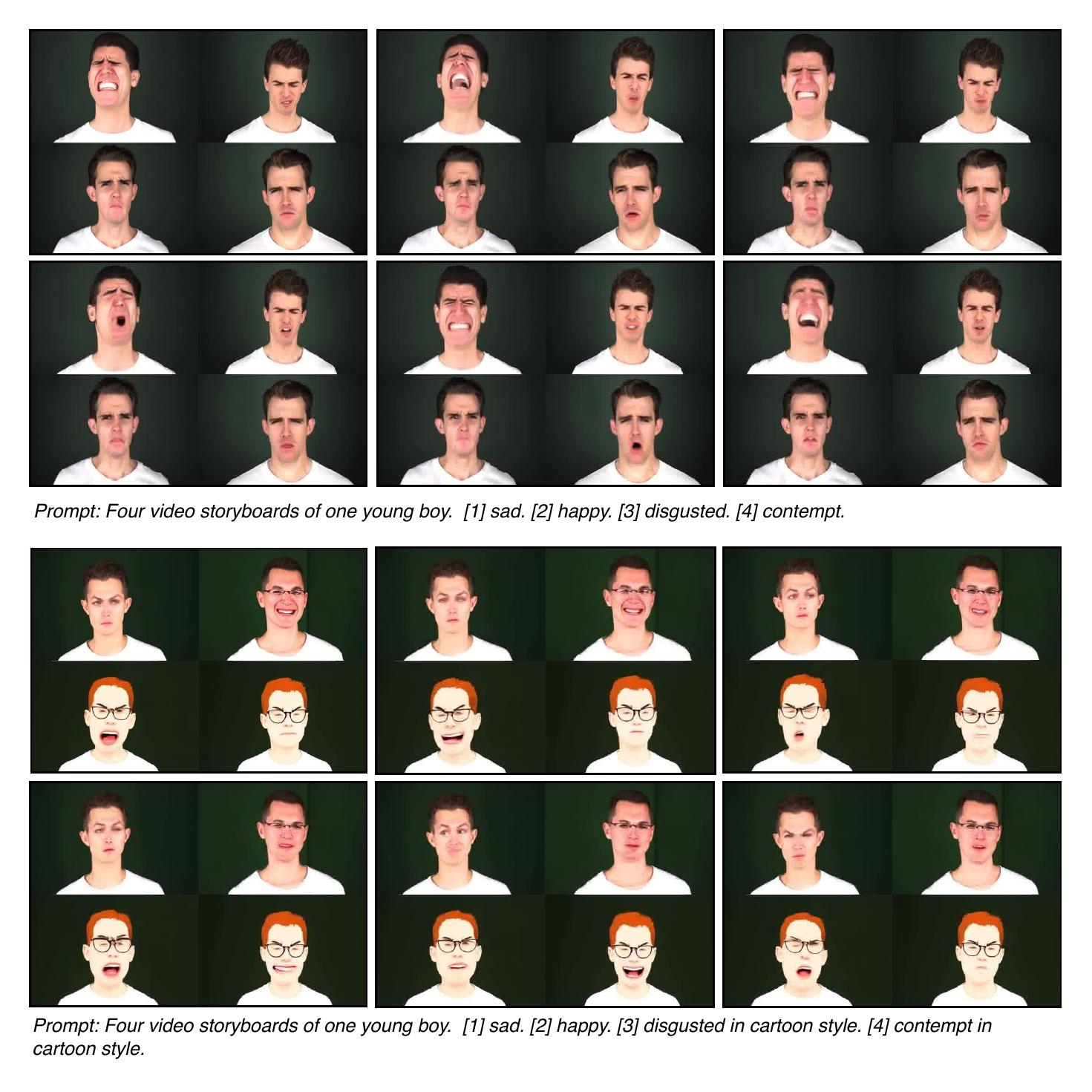}
   \caption{\textbf{Examples of in-context generalist for portrait photograph and style transfer tasks.}
   Four sub-videos are generated simultaneously within a single diffusion process being specifically tuned for the desired outcome. Consistent subject identities are preserved across all sub-videos within each set, as demonstrated in the accompanying figure.
   }
   \label{fig:case2} 
\end{figure*}

\begin{figure*}[htbp]
  \centering
   \includegraphics[width=0.99\linewidth]{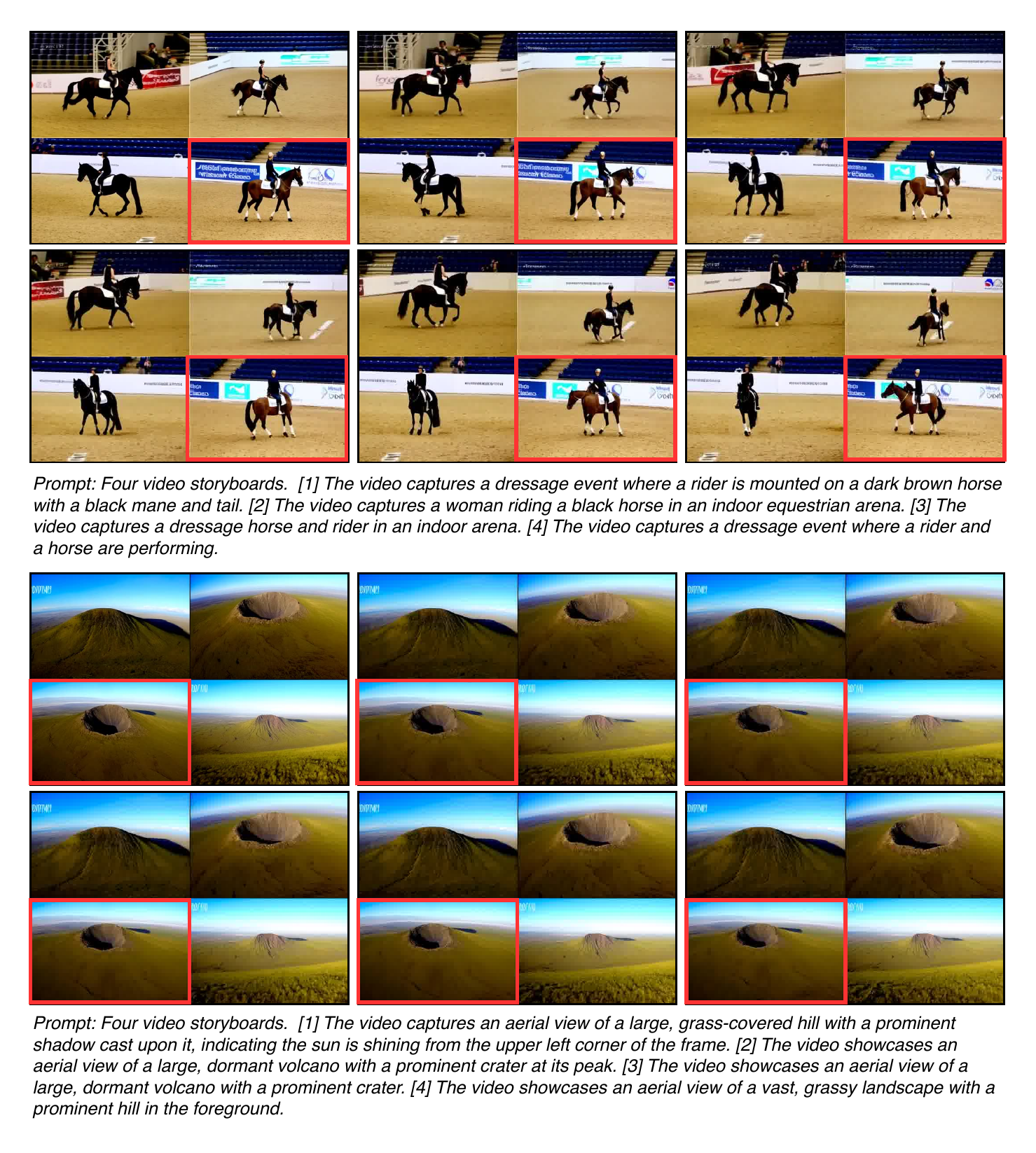}
   \caption{
   \textbf{Examples of in-context generalist applied to inpainting and outpainting tasks.}
   A sub-video, enclosed within a red box, is generated based on remaining video clips using a masking operation. This process ensures a consistent style and subjective maintained across all videos within each set.
   }
   \label{fig:case3} 
\end{figure*}

\subsection{Controllable Video Generation}

As video generation techniques advanced, early approaches to conditioning diffusion models with spatial controls primarily relied on concatenation \cite{rombach2022high, saharia2022palette}. The emergence of large-scale, pretrained text-to-image models \cite{esser2021taming, esser2024scaling} has harnessed sophisticated text embeddings \cite{radford2021learning, raffel2019exploring} processed through refined attention mechanisms, propelling research into more precise control over the generation process \cite{li2025controlnet, mou2024t2i, peng2024controlnext,fei2023gradient}. Notably, ControlNet \cite{zhang2023adding} facilitates fine-grained control over attributes such as depth, edges, and poses by embedding timestep-specific control features into the diffusion U-Net architecture. Additional control signals have also been explored, including camera motion \cite{motionctrl, he2024cameractrl, xu2024camco, feng2024i2vcontrol}, objective motion \cite{wang2024boximator, directavideo}, and trajectory control \cite{dragnuwa, wu2025draganything}, among others \cite{niu2024mofa, fang2024motioncharacter}.
\cite{zhang2024video} analyze the video in-context learning in autoregressive transformers.
Building on \cite{huang2024context}, we propose a unified framework that consolidates these diverse controls through concatenation, generating video content using both textual prompts and masking operation.

\section{Methodology}

\subsection{Task Formulation}

The generation of long-duration multi-scene videos is typically framed as the creation of a set of $n \geq 1$ videos, conditioned on another set of $m \geq 0$ videos and $(n+m)$ text prompts. This formulation also encompasses a wide variety of controllable video generation tasks, including video translation, style transfer, pose transfer, and subject-driven generation, as well as practical applications such as storyboard video generation. The relationships between the conditional and generated videos are implicitly steered through the individual video-specific prompts.

To achieve this goal, we use a slight modification to the conventional video diffusion transformers by utilizing a single, consolidated prompt template for the entire set of videos, not touching the network structure itself. This unified prompt begins with an overarching description of the entire video set, followed by specific prompts for each individual video. This design is compatible with existing text-to-video models and ensures that the overall description effectively communicates the task's intended objectives.

\subsection{In-Context Generation}

We begin with the premise that through large-scale video pretraining, video diffusion transformer inherently understand the cross-frame video representation \cite{huh2024platonic}, i.e., existing text-to-video models are able to possess some level of in-context generation capability across a range of tasks, albeit with varying quality. Leveraging this insight, we aim to activate the model's in-context generation potential through the use of meticulously curated, high-quality text-video sets. Simultaneously, we seek to maintain the integrity of the original text-to-video architecture, avoiding any structural modifications. Following \cite{huang2024context}, we propose the use of a single prompt that encompasses descriptions of multiple videos from one source, thereby enabling the desired consistent functionality within the existing framework.

Our proposed design generates a set of videos concurrently by concatenating them into a single video clip in tempera or spacial dimension during the training phase. The corresponding captions for these videos are integrated into a unified prompt template, which includes both an overarching description and specific guidance for each sub video clips. 
Once the video set is generated, we partition the large video into separate panels. Additionally, we apply Lora \cite{hu2021lora} to a selected subset of high-quality text-video data to enhance the capacity of models. 
To facilitate conditioning on an additional set of videos, we modify the inpainting scripts from image diffusion models\footnote{https://huggingface.co/docs/diffusers/using-diffusers/inpaint}. This training-free masking operation allows us to inpaint or outpaint the target video parts based on a partially masked video set, all of which are concatenated into the same large video.

\section{Experiments}

\subsection{Implementation Details}

We leverage CogvideoX-5b text-to-video model\footnote{https://huggingface.co/THUDM/CogVideoX-5b} as the foundation and specifically train an LoRA tailored to the unique requirements of specific tasks. We focus on a diverse set of practical applications, including multi-scene video generation, visual identity design, and portrait photography, among others. For each task, we curate a dataset of approximately 100 to 200 high-quality video sequences, sourced either from the internet or publicly available datasets. These video sets are then concatenated into composite videos, either along the spatial or temporal dimensions, and accompanied by detailed captions generated using Qwen2-VL\footnote{https://github.com/QwenLM/Qwen2-VL}.
The training process is carried out on 8 NVIDIA A100 GPUs, with 5000 optimization steps of 1e-5 learning rate, a batch size of 128, and a LoRA rank of 32. For inference, we adopt 50 sampling steps with a guidance scale set to 6, in accordance with the default settings of the CogVideoX-5b in Factory\footnote{https://github.com/a-r-r-o-w/cogvideox-factory}. For controllable video-conditional generation, we predict masked patches that facilitate inpainting or outpainting by utilizing surrounding contextual information at each frame.

\subsection{Results}

We finally provide qualitative results to illustrate the adaptability and performance of video diffusion transformers equipped with in-context learning capabilities across diverse tasks.

\paragraph{Reference-free Video-Set Generation.}
In this scenario, video sets are created exclusively from textual prompts, without reliance on any input videos. Illustrative examples spanning various applications, such as multi-scene composition, portrait photography, and style transfer, are depicted in Figures \ref{fig:case1} and \ref{fig:case2}. The proposed approach consistently produces outputs of exceptional quality.

\paragraph{Reference-based Video-Set Generation.}

In this configuration, video sets are synthesized using a combination of a textual prompt and a set of conditional videos, which must include at least one reference video. Masking techniques are applied to certain videos, enabling either inpainting or outpainting by leveraging information from the remaining frames. Results showcasing video-conditioned generation are provided in Figure \ref{fig:case3}.

\bibliographystyle{ieee}
\bibliography{main}

\end{document}